\pdfoutput=1

\documentclass[11pt]{article}

\usepackage[final]{acl}

\usepackage{times}
\usepackage{latexsym}

\usepackage[T1]{fontenc}

\usepackage[utf8]{inputenc}

\usepackage{microtype}

\usepackage{inconsolata}

\usepackage{graphicx}

\usepackage{hyperref}
\usepackage{amsmath} 
\usepackage{float}
\usepackage{comment}

\usepackage[framemethod=TikZ]{mdframed} 

\usepackage{enumitem}

\title{Revisiting Active Learning under (Human) Label Variation} 

\author{Cornelia Gruber$^{*1}$ 
        \quad Helen Alber$^{*1,2}$
        \quad Bernd Bischl$^{1,2}$\\  
     \textbf{Göran Kauermann}$^{1,2}$  
       \quad \textbf{Barbara Plank}$^{2,3}$
 \quad \textbf{Matthias Aßenmacher}$^{1,2}$\\
        $^1$ LMU Munich, Department of Statistics,  Germany \\ 
        $^2$ Munich Center for Machine Learning (MCML), Germany\\ 
        $^3$ LMU Munich, Center for Information and Language Processing (CIS), Germany\\ 
         \small{$^{*}$Equal contribution \quad
   \textbf{Correspondence:} \href{mailto:cornelia.gruber@lmu.de}{cornelia.gruber@lmu.de}, \href{mailto:helen.alber@stat.uni-muenchen.de}{helen.alber@stat.uni-muenchen.de}
 }
       }

\begin{document}
\maketitle

\begin{abstract}
Access to high-quality labeled data remains a limiting factor in applied supervised learning. While label variation (LV), i.e., differing labels for the same instance, is common, especially in natural language processing, annotation frameworks often still rest on the assumption of a single ground truth. This overlooks human label variation (HLV), the occurrence of plausible differences in annotations, as an informative signal. Similarly, active learning (AL), a popular approach to optimizing the use of limited annotation budgets in training ML models, often relies on at least one of several simplifying assumptions, which rarely hold in practice when acknowledging HLV. 
In this paper, we examine foundational assumptions about truth and label nature, highlighting the need to decompose observed LV into signal (e.g., HLV) and noise (e.g., annotation error). We survey how the AL and (H)LV communities have addressed---or neglected---these distinctions and propose a conceptual framework for incorporating HLV throughout the AL loop, including instance selection, annotator choice, and label representation. We further discuss the integration of large language models (LLM) as annotators. Our work aims to lay a conceptual foundation for HLV-aware active learning, better reflecting the complexities of real-world annotation.
\end{abstract}

\section{Introduction}
Prediction algorithms play a central role in many natural language processing (NLP) tasks, like hate speech detection \citep{basile2020ItsEndGold}, sentiment analysis \citep{kenyon-dean-etal-2018-sentiment}, or natural language inference (NLI; \citealp{pavlick2019InherentDisagreementsHuman}). For training such supervised machine learning (ML) models, a notable amount of labeled training data is necessary. However, acquiring high-quality labels is expensive as human crowd workers or, even more expensive, domain experts need to annotate the data. A popular scheme to efficiently guide the annotation process and allocate annotation budgets is active learning \citep[AL;][]{abney2007SemisupervisedLearningComputational, settles2009}. AL aims to maximize the expected predictive performance of the resulting model while minimizing the required number of annotations; often done by iterating the following three steps: (1) Training the ML model on available labeled data, based on measure $m$. (2) Selecting new instances for labeling from a pool of unlabeled data, usually based on an acquisition function. (3) Labeling these with an oracle. 
Those steps, which are repeated until the available annotation budget is depleted or the model has reached its target accuracy, rest on the following assumptions: 
\begin{enumerate}
    \item[\textbf{A1}] There exists a single ground truth label per instance.
    \item[\textbf{A2}] The oracle provides the ground truth labels without any noise.
    \item[\textbf{A3}] The annotation difficulty or cost is equal for all instances.
\end{enumerate}

\begin{figure}[t]
    \centering
    \includegraphics[width=1\linewidth]{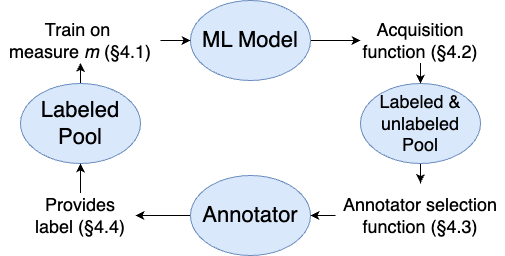}
    \caption{The traditional AL loop with possible adaptations in different steps, leading to generalized label variation aware AL}
    \label{fig:AL_loop}
    \vspace*{-0.2cm}
\end{figure}

\noindent Equal annotation cost is not, strictly speaking, a critical assumption for AL, but is becoming increasingly important to consider. However, in NLP, those assumptions often are not or cannot be fulfilled. Especially in the presence of human label variation (HLV), i.e., differences in human annotations that are plausible variability due to subjectivity or ambiguity and explicitly no sign of error (\citealp{plank-2022-problem}; cf. \S\ref{sec:relwork}), even the existence of such an omniscient oracle is questionable.

When we move away from these assumptions and acknowledge HLV, the AL loop is extended: an annotator selection function is introduced to choose among multiple annotators with varying perspectives or expertise, rather than assuming a single infallible oracle (cf. \autoref{fig:AL_loop}).

\paragraph{Contributions} In this work, we examine the consequences for the AL cycle when its conventional assumptions, i.e., A1 -- A3, are violated due to plausible variation in labels, often coined HLV. We begin by discussing foundational assumptions about truth in annotation (\S\ref{sec:assump}), laying out different perspectives on label nature and emphasizing the need for a signal-noise decomposition of label variation (LV) into plausible variation (e.g., HLV) and noise. In what follows, we provide an overarching survey of the literature of both fields, i.e., (H)LV and AL, that reveals an emerging line of research integrating aspects of (H)LV into AL (\S\ref{sec:relwork}), but simultaneously also uncovers shortcomings and misunderstandings between the fields. We then identify and categorize the adaptations required in the AL loop (\S\ref{sec:adaptations}), including modifications to the annotator selection function and considerations for incorporating LLMs. Altogether, we offer a holistic perspective on AL in the presence of (H)LV, aiming to establish a more structured ground for discussion and future empirical investigation by bridging ongoing debates across NLP, empirical ML, statistics, and philosophy.

\section{Assumptions about Truth in Annotation} 
\label{sec:assump}

When observing LV in human annotations, it is important to recognize that this variation may arise from both error and HLV \citep{weber-genzel-etal-2024-varierr}, which can be present simultaneously. Throughout this work, we use LV to refer to the observed differences in annotation, which can be decomposed into signal, such as HLV, and noise, such as actual annotation error. Reflecting on the underlying assumptions about the true labels is crucial, as it helps to distinguish between these sources of LV, or, in other words, aids the ``interpretation of any observed annotator disagreement'' \citep[][p. 3]{rottger-etal-2022-two}.

\paragraph{Task Dependence and Subjectivity}
The extent to which observed LV is attributed to HLV is often judged based on the (assumed) subjectivity of the task~\cite{basile-etal-2021-need}. In domains such as specific image classification tasks in computer vision (e.g., distinguishing between images of cats or dogs), lower levels of HLV may be expected, as the real-world categories constituting the data-generating process (e.g., actual cats or dogs) are typically less subjective. In such cases, higher shares of the observed label variation may be attributable to various types of errors, such as issues arising from imprecise measurement, the compression of real-world information into data, or noise, e.g., introduced during data collection like blurriness in images \citep{gruber2025SourcesUncertaintySupervised}, rather than to HLV. 

The notion of inferring task subjectivity from observed LV introduces a certain circularity: LV is intuitively taken as evidence of subjectivity, while assumptions about subjectivity, in turn, inform how much of the variation is attributed to HLV. A more thorough discussion and a systematic approach to operationalizing subjectivity lie beyond the scope of this work, but appear essential when aiming to disentangle signal and noise in observed LV.

\paragraph{Worldviews and Nature of Truth}
Many NLP tasks, as well as certain computer vision tasks \citep[e.g., image segmentation in medicine;][]{zhang_disentangling_2020}, are assumed to involve a higher degree of subjectivity. Particularly when addressing such tasks, different underlying philosophical assumptions on the nature of truth and the closely related nature of reality can lead to varying methodological implications. For example, adopting a monistic worldview---drawing on the discussion of monism by \citet{russell_iinature_1907}---may involve the assumption of a single underlying reality, with different annotations merely being different perspectives on it. In this context, no observed annotation could be fully true or false, and taking individual annotations into account as a distribution on the instance level may be a reasonable approach.

\paragraph{Label Non-Determinism and Levels} 
While a comprehensive philosophical discussion is beyond the scope of this work, one more practical consideration warrants a brief mention. Whether label variation is viewed from the annotator’s perspective (annotator level) or the instance’s perspective (instance level) can help clarify certain complexities. For example, on the annotator level, label non-determinism, defined as a probabilistic mapping between a real-world instance and a set of labels, can vary in degree between both subjective and less subjective cases, and may even include label-deterministic subjective settings. In contrast, on the instance level, greater subjectivity inherently results in more label non-determinism. Ambiguity, here clearly distinguished from subjectivity, is linked to higher label non-determinism at both levels. While factors like these---label non-determinism, subjectivity, ambiguity, and annotator level vs. instance level---can, in principle, be treated separately, we assume substantial dependencies between them. For instance, even at the annotator level, tasks that are assumed to be more subjective may be likely prone to exhibiting a greater degree of label non-determinism. 

\begin{figure*}[ht]
    \vspace*{-10pt}
    \centering
    \includegraphics[width=.7\linewidth]{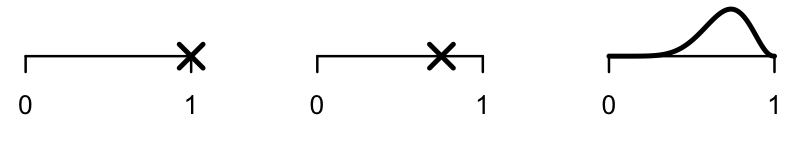} \\
    \includegraphics[width=.7\linewidth]{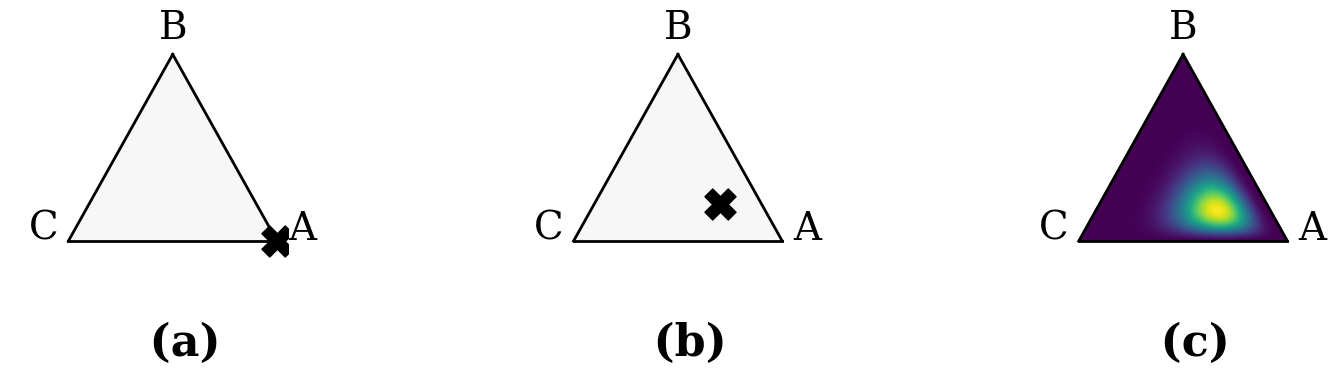}
    \caption[Types of labels]{Types of labels visualized. Each label  \(y_i \) is a realization of a random variable \(Y \).\\\hspace{\textwidth}
    Top row: binary classes; bottom row: three classes. \\\hspace{\textwidth}
    \textbf{(a)} Discrete label: \(y_i = 1 \) with \( Y \sim \text{Bin}(1, p) \) (top) and \(y_i = [1, 0, 0] \) with \( Y \sim \text{Multinom}(1, \pmb{p}) \) (bottom), \\\hspace{\textwidth}
    \textbf{(b)} probability label: \(y_i = 0.75 \) with \( Y \sim \text{Beta}(\alpha, \beta) \) (top) and \(y_i= [0.6, 0.2, 0.2] \) with \( Y \sim \text{Dir}(\pmb{\alpha}) \) (bottom), \\\hspace{\textwidth}
    \textbf{(c)} distributional label: \(y_i = \text{Beta}(8, 3.5) \) (top) and \(y_i = \text{Dir}(8, 3, 4) \) (bottom) with a hierarchical model for the distribution of \(Y\) with priors on the parameters of the respective distributions. In the bottom row, ternary plots visualize the relative proportions of three classes as positions within a triangle. Each cross represents a single label, with its location indicating the class composition: the closer a point is to a corner, the higher the class proportion.}
    \label{fig:types_of_labels}
\end{figure*}

\paragraph{Types of Label Nature}
Approaching the discussion from a more applied perspective, we provide an overview of possible types of labels: 
(a) discrete class label (also known as ``hard label''), (b) label as probability for discrete classes \citep[sometimes referred to as ``soft label'',][or ``human judgment distribution'']{uma2021LearningDisagreementSurvey}, and (c) label as continuous distribution for underlying fixed number of classes (cf. \autoref{fig:types_of_labels}). Note, that while the illustration depicts only scenarios with $k \in \{2, 3\}$ classes for simplicity, this schema is generally also applicable to settings with $k > 3$ classes.
When viewing the annotation process from a statistical perspective, i.e. making assumptions about the data generating process, each label \(y_{i} \) can be regarded as a realization of a random variable \(Y \). For discrete labels (a), an example in the binary setting is \(y_{i} = 1 \), with \( Y \sim \text{Bin}(1, p) \); in the ternary case, i.e., three classes, an example is \(y_{i} = [1, 0, 0] \), with \( Y \sim \text{Multinom}(1, \pmb{p}) \), \(\pmb{p} = (p_{A}, p_{B}, p_{C})\).
Moving to probability labels (b), the label itself represents a probabilistic belief over class membership. For instance, \(y_{i} = 0.75 \) may arise from \( Y \sim \text{Beta}(\alpha, \beta) \), and \(y_{i} = [0.6, 0.2, 0.2] \) may be a realization from \( Y \sim \text{Dir}(\pmb{\alpha}) \), \(\pmb{\alpha} = (\alpha_{A}, \alpha_{B}, \alpha_{C})\).
Finally, in the case of distribution labels (c), \(y_{i} \) takes the form of a full probability distribution---for example, \(y_{i} = \text{Beta}(8, 3.5) \) in the binary case or \(y_{i} = \text{Dir}(8, 3, 4) \) in the ternary case. Here, the label \(y_{i} \) is itself a distribution over class probabilities. The distribution of \(Y\) is modeled hierarchically by placing priors on the parameters of this distribution, e.g., on \(\alpha \), \(\beta \) in the Beta case or on \(\pmb{\alpha} \) in the Dirichlet case \citep{hechinger2024HumanintheloopLabelEmbeddings}.

We here challenge the common assumption of the first type (discrete class labels, sometimes also referred to as ``single ground truths'') by proposing the consideration of the latter two types, both as assumed true labels and requested annotations. The third label type appears to be the least studied of the ones listed; however, some work in uncertainty quantification has begun to explore different label representations \citep{bengs2022pitfalls, wimmer2023QuantifyingAleatoricEpistemic, sale2024SecondorderUncertaintyQuantification, hechinger2024HumanintheloopLabelEmbeddings}.

\noindent In practice, a discrepancy can occur between the type of label assigned by the annotator and the assumed nature of the true label. This mismatch is especially likely when true labels are assumed to be continuous distributions over classes (cf. case (c) in \autoref{fig:types_of_labels}), as human annotators are not inherently equipped to give non-discrete annotations (cf.\ \S\ref{oracle} for further discussion of the ``oracle'' in the AL cycle). This discrepancy introduces an irreducible uncertainty and may result in the interpretation that the observed label variation does not necessarily equate to HLV. This again emphasizes the importance of distinguishing between assumptions about the true labels and assumptions that may be required for practical reasons during annotation and the AL loop.

\section{Views on Label Variation and Active Learning}
\label{sec:relwork}

In what follows, we first review and attempt to categorize literature from multiple research domains that addresses label variation in data annotation---a phenomenon discussed under various terminologies, reflecting different theoretical perspectives and interpretations. We then examine how the field of AL has responded to, incorporated, or overlooked these diverse understandings of LV in its methodological developments. 

\subsection{Label Variation}
Supervised ML depends fundamentally on annotated data, making the quality and nature of labels a central part of the learning process. The phenomenon of LV, i.e., the occurrence of differing annotations for the same instance, both between and within annotators, is not limited to subjective tasks but has been found across a wide range of applications. In NLP, this includes tasks such as sentiment analysis \citep{kenyon-dean-etal-2018-sentiment}, hate speech detection \citep{basile2020ItsEndGold}, veridicality judgments \citep{de_marneffe_did_2012}, argumentation mining \citep{trautmann2020FineGrainedArgumentUnit}, natural language inference \citep{pavlick2019InherentDisagreementsHuman}, and even tasks traditionally considered ``objective'' like part-of-speech tagging \citep{plank-etal-2014-linguistically}, word sense disambiguation \citep{passonneau2012MultiplicityWordSense}, semantic role labeling \citep{dumitrache-etal-2019-crowdsourced}, and named entity recognition \citep{inel2017HarnessingDiversityCrowds}. Similar variation has also been observed in computer vision tasks like medical image classification and object identification \citep{uma2021LearningDisagreementSurvey}, or remote sensing \citep{hechinger2024CategorisingWorldLocal}, where annotator disagreement arises from ambiguity and subjectivity in visual interpretation. While most existing works listed treat this as \textit{either} signal or noise, we refrain from exclusively assigning observed label variation to either category in the first place.

\paragraph{Mitigating Label Variation}
The assumption of a single ground truth label has long dominated ML practice, as reflected in foundational ML literature \citep{mitchell_machine_1997, hastie2009ElementsStatisticalLearning, goodfellow2016DeepLearning}. Within this framework, LV is typically regarded as erroneous and to be minimized or corrected \citep{alm2011subjective, aroyo2015truth} with \citet{cabitza2023PerspectivistTurnGround}, for example, documenting widespread practices of ``disagreement removal''.

\paragraph{Treating ``Hard'' Cases}
Moving beyond the traditional view of LV, early work has begun to explore LV as a potential source of information. Exemplary, \citet{reidsma_exploiting_2008} advocate for analyzing patterns of disagreement, providing an overview of the various factors that may underlie annotator disagreement. However, this line of work uses information from LV to steer ML models away from ``hard'' cases (i.e., items with high LV), by, e.g., enabling classifiers to abstain from making predictions. \citet{plank-etal-2014-learning} propose incorporating inter-annotator agreement measures into a cost-sensitive loss function, thereby explicitly integrating LV into the learning process as a signal of uncertainty. The following approaches seek to embrace LV more directly by explicitly modeling it, for instance, through adjustments to the nature and interpretation of the labels.

\paragraph{Human Label Variation}
There are two main bodies of literature relevant to this work addressing differences in human annotations: one that predominantly uses the term \textit{variation} and another that refers to \textit{disagreement}. We adopt the terminology of \citet{plank-2022-problem}, who introduced the notion of HLV to conceptualize such differences as plausible and meaningful variations rather than as annotation errors. This perspective has been particularly motivated by developments in NLP, where subjectivity, leading to HLV, is recognized as an inherent property of many language-related tasks \citep{alm2011subjective}. This framework further aligns with the concept of \textit{perspectivism} introduced by \citet{cabitza2023PerspectivistTurnGround}, which emphasizes that, rather than seeking a single ground truth, collecting multiple labels offers a way to sample the range of perceptions, opinions, and judgments present in a population.

The related body of literature that adopts the term disagreement rather than variation is more heterogeneous in its interpretation and evaluation of annotation differences. While some contributions view such disagreement as plausible or informative \citep{uma2021LearningDisagreementSurvey}, others primarily treat it as a source of noise or error \citep{beigman_klebanov_annotator_2009}. Throughout this section, we review work from both terminological traditions.

\paragraph{Distributional Labels}
Several contributions have moved beyond discrete labels by aggregating multiple annotations into distributional labels (cf.\ \autoref{fig:types_of_labels} for the different types of label nature), aligning with a (strong) perspectivist stance. \citet{de_marneffe_did_2012} frame veridicality assessment as a ``distribution-prediction task'', using judgments from 10 annotators per instance. Similarly, \citet{aroyo2015truth} view disagreement as a signal and introduce the ``Crowd Truth'' framework, which incorporates distributional labels through annotation aggregation and addresses factors like the design of annotation guidelines and differing annotator expertise. 
In computer vision, \citet{Peterson2019HumanUM} show that training convolutional neural networks on soft labels derived from multiple annotators improves generalization under distributional shift. For NLI, \citet{pavlick2019InherentDisagreementsHuman} use slider-based annotations to capture uncertainty and argue for models that predict distributions over judgments. More recently, \citet{chen-yu-and-samuel-r-bowman-2022-clean} and \citet{gruber-etal-2024-labels} investigated whether to prioritize more annotators per instance or more annotated instances when working on the distributional level via label aggregation.

However, these contributions primarily address HLV by aggregating multiple annotations per instance, thereby treating distributional labels as post-hoc constructions rather than as \textit{distributional by nature}---as in case (c) above---i.e., labels deliberately designed from the outset to capture uncertainty directly as a characteristic of the label.

\paragraph{Decomposing Label Variation}
Furthermore, the above contributions tend to conflate LV with HLV, overlooking the simultaneous presence of \textit{both} noise and signal within LV. Incorporating this conceptual distinction, \citet{palomaki2018case} highlight the need to distinguish between actual annotation errors and ``disagreement that falls within the acceptable range'', introducing the concept of \textit{acceptable variation}, which may differ across subsets of instances and has direct implications for task design. \citet{weber-genzel-etal-2024-varierr} extend this conceptual distinction to NLI. They address the challenge of identifying annotation error by incorporating validated annotator \textit{labels with explanations} through a second round of validity judgments, rather than relying on post-hoc interpretation alone. This builds on earlier work by \citet{jiang_ecologically_2023}, who identify the phenomenon of \textit{within-label variation}, where, even when the same label is assigned, annotators may vary in their explanations.

Data annotation remains a labor-intensive and complex process, particularly when aiming to analyze or leverage a signal–noise decomposition of LV. The following section, therefore, turns to the field of active learning, which focuses on strategies for optimizing annotation budgets and minimizing annotation effort.

\subsection{Active Learning}
Active Learning has been a vivid field of research for over 30 years \citep{seung1992QueryCommittee, lewis1994HeterogeneousUncertaintySampling, settles2009, aggarwal2014ActiveLearningSurvey}.
\citet{settles_theories_2011} already discussed practical issues arising in active learning, including querying in batches, noisy oracles, and variable labeling costs.
\citet{zhang2022SurveyActiveLearning} provide a survey on AL for NLP, while \citet{rauch2023ActiveGLAEBenchmarkDeep} propose a tailored NLP benchmark for AL.


\paragraph{Annotation Costs and Quality}
The true costs of annotation are explored in \citet{margineantu2005ActiveCostsensitiveLearning, settles2008ActiveLearningReal, xie-etal-2018-cost, krishnamurthy2019ActiveLearningCostSensitive}, challenging assumption \textbf{A3} (\textit{``The annotation difficulty or cost is equal for all instances.''}) by modeling variation in annotation effort. \citet{gao2020CostAccuracyAwareAdaptive} extend this by accounting for annotators with varying accuracies, while \citet{donmez2008ProactiveLearningCostsensitive} acknowledge that even oracles might be incorrect depending on task difficulty, both relaxing \textbf{A2} (\textit{``The oracle provides the ground truth labels without any noise.''}). Furthermore, \citet{zhang2015ActiveLearningWeak} and \citet{chakraborty2020AskingRightQuestions} incorporate both low-cost and expert annotators by assuming a trade-off between cost and label quality. However, these approaches still assume a single ground truth label per instance (reliance on \textbf{A1}; \textit{``There exists a single ground truth label per instance.''}) and treat label variation as noise. 


In contrast, we highlight the underexplored setting where (H)LV is inherent and may carry an informative signal, arguing that its integration into the AL framework requires rethinking core components such as acquisition and annotation strategies.

\paragraph{Relabeling}
Relabeling, i.e., collecting additional annotations for previously labeled instances to reduce noise or correct errors, is explored in \citet{chen-yu-and-samuel-r-bowman-2022-clean, goh2023ActiveLabActiveLearning, lin2016ReActiveLearningActive}. These approaches implicitly challenge assumptions \textbf{A2} and \textbf{A3} by acknowledging annotation errors and varying difficulty. However, they treat disagreement as an error rather than a potentially meaningful signal. 

\paragraph{HLV-aware AL}


A few recent studies have begun to explore how AL can be adapted to account for HLV.
\citet{wang-plank-2023-actor} and \citet{van-der-meer-etal-2024-annotator} suggest strategies to choose \textit{which} human annotator should label an instance.
Furthermore, \citet{baumler_which_2023} suggest aligning model
uncertainty with annotator uncertainty.
While these works offer valuable insights, they address specific assumptions or propose targeted adaptations to the AL process. In \S\ref{sec:adaptations}, we build on these efforts by systematically analyzing their contributions and organizing them into a broader framework. There, we formalize and categorize key adaptations required for making AL effective in the presence of HLV, and point to open challenges and directions for future research.

\section{The Active Learning Loop Revisited}
\label{sec:adaptations}

In the following, we discuss the consequences of the assumptions about truth in annotation and the nature of the labels (\S\ref{sec:assump}) on each of the steps of the AL loop (as visualized in \autoref{fig:AL_loop}).

\subsection{Training Measure}
Traditional AL assumes a single ground truth label provided by an oracle. This aligns naturally with classic supervised ML, where models are optimized based on hard-label measures like Bernoulli loss or cross-entropy. However, in cases where label variation is not due to error but comes from plausible causes, different soft-label measures are necessary. In such cases, alternative loss measures based on label distributions, such as Kullback-Leibler (KL) divergence \citep{koller2024GoingOneHotEncoding}, Jensen-Shannon divergence, or label embeddings \citep{schweden2025CanUncertaintyQuantification} have been proposed. \citet{baumler_which_2023} offer solutions by comparing the predicted and observed label distribution, thus directly optimizing for a trustworthy representation of LV. \\

\begin{mdframed}[roundcorner=2pt,linewidth=2pt]
\textbf{(C1) Consequence:} In the presence of HLV, distributional measures must be used for optimizing and evaluating the classifier.
\end{mdframed} 

\subsection{Acquisition Function}
The acquisition function ranks all unlabeled instances by their usefulness if they were to be labeled. The oracle then provides labels to the most instructive cases. Traditional AL~\cite{zhang2022SurveyActiveLearning} uses querying strategies based on either \textit{informativeness} or \textit{representativeness}, or \textit{hybrid} approaches \citep{Ash2020Deep}. 
Informative querying often uses uncertainty sampling \citep{lewis1994SequentialAlgorithmTraining}, where the samples with the highest predicted label entropy get labeled first, thus the ones with the highest uncertainty. However, with HLV, high entropy can also be integral to the task, and thus not necessarily a sign of uncertainty. This shows that classic entropy sampling is not suitable for AL in the presence of HLV. 
Representativeness sampling favors samples that represent the unlabeled pool well. However, classical representativeness sampling ignores the option of labeling some instances multiple times to represent HLV properly and is thus also unsuitable for HLV. Further, defining representativeness in distributions is not trivial.   
One option to take HLV into account is to precede the AL loop by training a prediction model for annotator disagreement (entropy) and then changing the acquisition function to query samples where the predicted annotator entropy and model entropy diverge the most \citep{baumler_which_2023}.\\

\begin{mdframed}[roundcorner=2pt,linewidth=2pt]
\textbf{(C2) Consequence:} In the presence of HLV, classical informativeness or representativeness sampling are unsuitable, as they ignore the option of labeling instances multiple times and fail to process distributional labels. 
\end{mdframed}

\subsection{Annotator Selection Function}
The assumption of having an oracle providing the single ground truth label is not suitable in subjective tasks, where the distribution of human opinions is of interest, or other tasks with high (assumed) HLV. Therefore, an additional step in the AL cycle needs to be considered: the selection of annotators. In many crowd worker settings, it is possible to inquire about labels from a specific annotator.
Extending this thought, different ``types'' of annotators could be queried, e.g., not only human workers but also large language models (LLM). This is also known as ``pre-annotation'' \citep{zhang2022SurveyActiveLearning} in the pre-LLM era, and analogously as  ``LLM-as-annotator'' \citep[or ``LLM-as-a-judge'';][]{zheng2023judging,wu2024metarewardinglanguagemodelsselfimproving} today, where the idea is that a model's predictions are given to human annotators to confirm or adjust. 
Consequently, an overarching annotator selection strategy needs to evaluate whether a language model or a human shall provide the label, and whether a specific annotator (e.g., representing a minority) or a specific LLM could provide the label.
Recent work has extended the AL framework to include not only sample selection but also annotator selection. \citet{wang-plank-2023-actor} introduce a multi-head model that jointly selects the most informative instance and the most suitable annotator. In contrast, \citet{van-der-meer-etal-2024-annotator} focus on ensuring representativeness and diversity in annotator selection, proposing a strategy that balances labeler perspectives to reflect the underlying population of interpretations better. 
The idea of using LLMs as annotators is pursued in
\citet{zhang-etal-2023-llmaaa} and \citet{bansal2023LargeLanguageModels}. \\

\begin{mdframed}[roundcorner=2pt,linewidth=2pt]
\textbf{(C3) Consequence:} In the presence of HLV it matters \textit{who} provides the label. An annotator acquisition function must decide not only whether to query a human or a language model, but also \textit{which} specific annotator or model to select.
\end{mdframed}

\subsection{Quality of Label and Uncertainty}
\label{oracle}
The quality of annotators is an important area of research in NLP, which becomes increasingly meaningful when diversity in annotations is present (or required; \citealp{sorensen_position_2024}) and label noise cannot be easily separated from the plausible share of label variation. 
Currently, most work either assumes variation is noise \citep{zhang2015ActiveLearningImbalanced, zhao2011IncrementalRelabelingActive, goh2023ActiveLabActiveLearning} or all variation in labels represents true HLV \citep{wang-plank-2023-actor,van-der-meer-etal-2024-annotator}. 
Particularly, when the ground truth label is a distribution and multiple annotators provide labels, detecting annotation noise in HLV is a complex endeavor \citep{weber-genzel-etal-2024-varierr}. 
Now, when not only different humans annotate the data, but samples can also be processed by LLMs, assessing the label quality is non-trivial either \citep{ni2025CanLargeLanguage}. 
Also, in the process of labeling, human annotators usually provide a single label, while an LLM could directly provide distributions \citep{chen-etal-2024-seeing, pavlovic-poesio-2024-effectiveness}. This makes LLMs as annotators especially attractive in the presence of HLV and for providing labels for case (c) depicted in \autoref{fig:types_of_labels}.\\

\begin{mdframed}[roundcorner=2pt,linewidth=2pt]
\textbf{(C4) Consequence:} In the presence of HLV, it is non-trivial to distinguish true label variation from noise, especially when labels can be sourced from both humans and language models, each with differing capabilities and output formats. 
\end{mdframed}

\section{Conclusion}

In this work, we provide an overview of the crucial connection between the fields of (human) label variation and active learning. Our comprehensive review of the existing literature in the individual fields helps building bridges between different, but connected, streamlines of research, paving the way for the identification of critical aspects to consider in the AL loop in the presence of HLV. Our critical assessment of these aspects aims to further point out potential avenues for future research to deal with them in a more nuanced and reflective manner. In doing so, we uncover several crucial assumptions about labels which are often \textit{implicitly} made in traditional AL. However, we argue that they need to be made \textit{explicit}. While providing a unified and implemented solution to the discussed problems is beyond the scope of the paper, we still hope to contribute to ongoing research debates on (H)LV by providing a fresh perspective from a different angle on existing problems and encourage new work addressing label-variation-aware active learning.


\section*{Limitations}
While this work provides a structured discussion on active learning in the presence of human label variation, several limitations remain. The philosophical discussion on annotation truth is a conceptual suggestion rather than a prescriptive framework. For example, we do not address annotation tasks where it is assumed that \textit{no} ground truth exists, or discuss other frameworks like imprecise probabilities for representing human label variation. 
Moreover, not all discussed adaptations are implemented in AL pipelines yet, requiring empirical validation.
Additionally, we do not explore alternative methods for gathering human annotations that may better accommodate HLV in detail.
Lastly, the reliability of ``LLM-as-annotator'' remains an open question. While LLMs can reduce costs and provide label distributions, their biases and lack of accountability pose challenges.

\section*{Acknowledgments}

CG is supported by the DAAD programme Konrad Zuse Schools of Excellence in Artificial Intelligence, sponsored by the Federal Ministry of Education and Research. HA and MA are funded by the Deutsche Forschungsgemeinschaft (DFG, German Research Foundation) under the National Research Data Infrastructure – NFDI 27/1 - 460037581.

\bibliography{anthology, references-cg, references}



\end{document}